\documentclass{article} 
\usepackage{iclr2027_conference,times}

\usepackage{amsmath,amsfonts,bm}

\def\eqref#1{equation~\ref{#1}}

\def\1{\bm{1}}

\DeclareMathAlphabet{\mathsfit}{\encodingdefault}{\sfdefault}{m}{sl}
\SetMathAlphabet{\mathsfit}{bold}{\encodingdefault}{\sfdefault}{bx}{n}

\usepackage{hyperref}
\hypersetup{pdfborder={0 0 0}}
\usepackage{url}
\usepackage{algorithm}
\usepackage{algpseudocode}
\usepackage{booktabs}
\usepackage{graphicx} 
\usepackage{multirow}
\usepackage{subcaption}
\usepackage{wrapfig}
\usepackage{placeins}

\title{Accelerating Visual On-Policy Distillation with Batched Speculative Jacobi Rollouts}

\author{Bingqi Shan, Zhehao Yu, Kenhong Lin, Baoquan Zhang \\
Shenzhen Key Laboratory of Internet Information Collaboration\\
Harbin Institute of Technology, Shenzhen\\
Shenzhen, China \\
\texttt{\{23s051004,yuzhehao,linkenghong\}@stu.hit.edu.cn} \\
\texttt{baoquanzhang@yeah.net}
}

\iclrfinalcopy
\begin{document}

\maketitle
\fancyhead{}
\lhead{Preprint}

\begin{abstract}
Visual on-policy distillation (OPD) improves the training of compact visual autoregressive models by learning from trajectories generated by the current student. However, these online rollouts are still produced token by token with autoregressive decoding, which adds substantial cost to every on-policy training step. Speculative Jacobi Decoding (SJD) provides an alternative because it can process multiple tokens in parallel without an auxiliary draft model, but the original method is designed for single-sequence inference. We introduce HB-SJD, a batched SJD rollout backend for visual OPD. HB-SJD allows each image to advance independently according to its own decoding progress, while images at different sequence positions are still verified in batched model forwards. As images finish, HB-SJD switches between Full and Compact execution to reduce the cost of later rollout rounds. HB-SJD only replaces the student rollout backend and leaves the teacher, distillation objective, and optimization procedure unchanged. Experiments with LlamaGen show that HB-SJD substantially reduces rollout and end-to-end training time while preserving the generation quality of the distilled student.
\end{abstract}

\section{Introduction}

Visual autoregressive (AR) models have shown strong generation quality and scalability in both next-token and next-scale image generation \citep{sun2024autoregressive,tian2024visual}. As these models grow, knowledge distillation provides a practical way to transfer their capabilities to smaller students \citep{hinton2015distilling,kim2016sequence}. Recent work further extends distillation to generative models by matching sequence-level behavior and training on student-generated trajectories \citep{gu2024minillm,agarwal2024policy}. However, conventional distillation mainly trains the student on data or teacher trajectories, while the student must condition on its own previous outputs during inference. On-policy distillation reduces this mismatch by letting the current student generate training trajectories and using the teacher to supervise the resulting student-visited states \citep{agarwal2024policy,peruzzo2026knowledge}. This better matches the student's inference behavior, but it also requires online student rollouts during training, adding substantial cost to every on-policy update.

\begin{figure}[t]
    \centering

    \begin{subfigure}[t]{0.525\linewidth}
        \centering
        \scalebox{1}{%
            \includegraphics[
                width=\linewidth,
                trim=5pt 5pt 5pt 5pt,
                clip
            ]{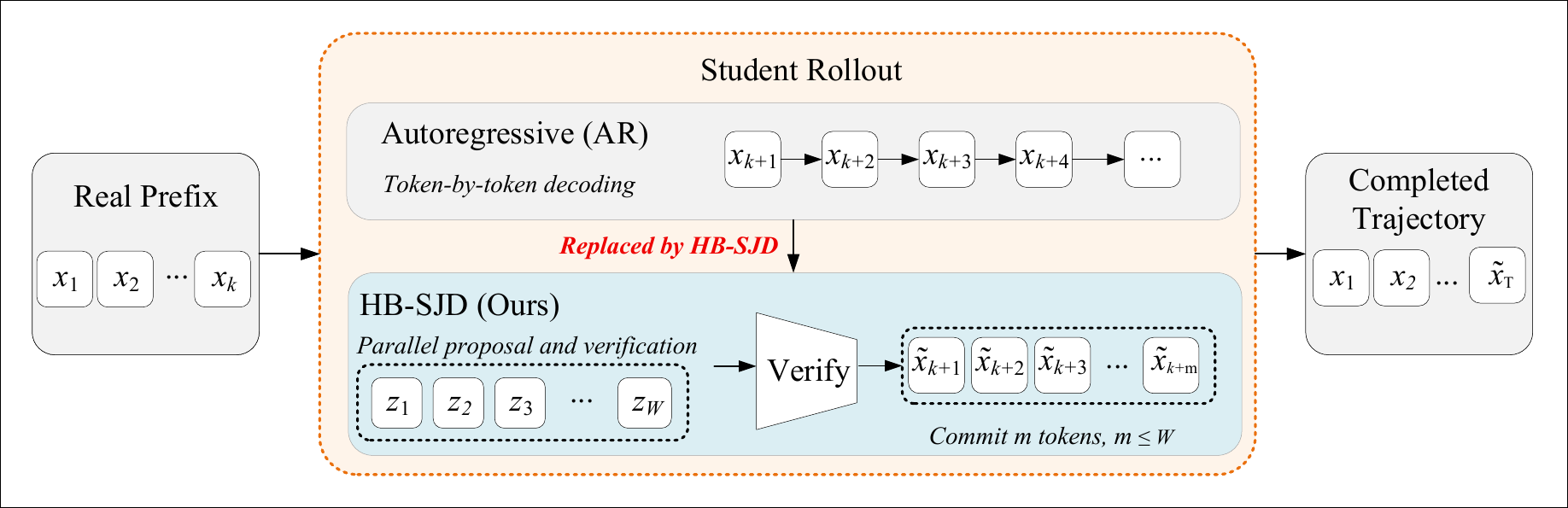}%
        }
        \caption{Student rollout replacement in visual OPD.}
        \label{fig:intro_a}
    \end{subfigure}%
    \hfill
    \begin{subfigure}[t]{0.455\linewidth}
        \centering
        \scalebox{1}{%
            \includegraphics[
                width=\linewidth,
                trim=5pt 5pt 5pt 5pt,
                clip
            ]{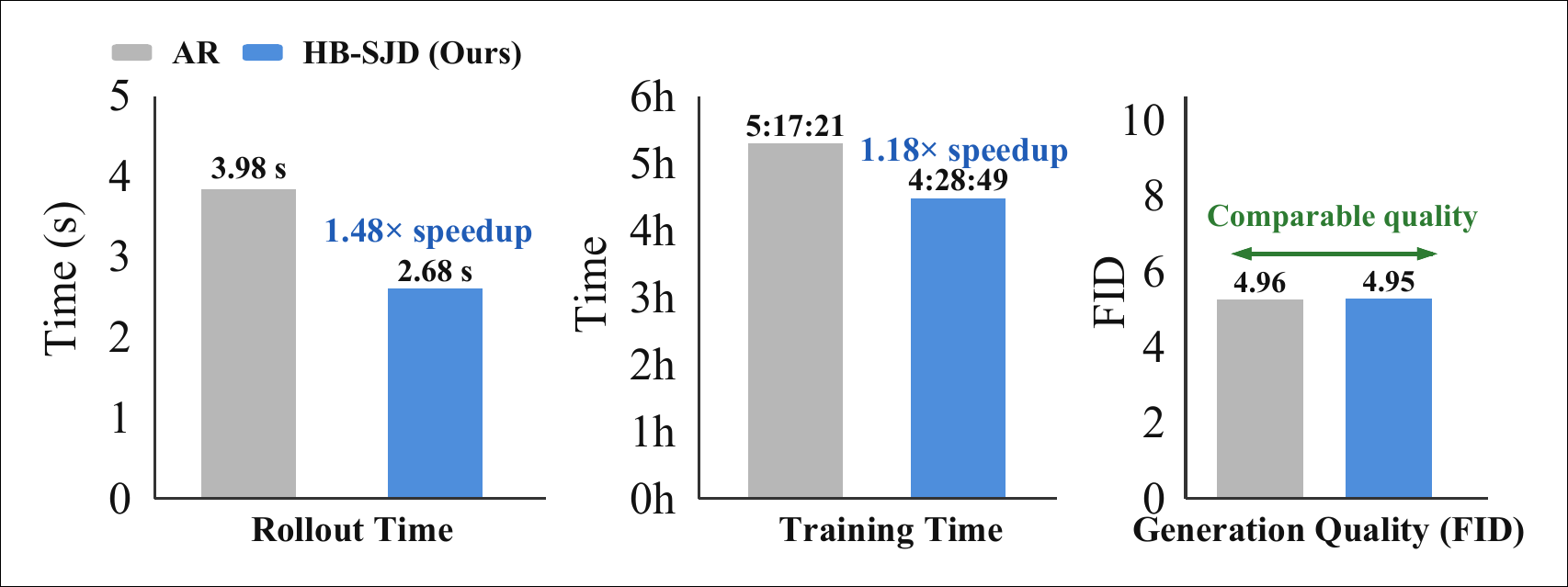}%
        }
        \caption{Cost and performance comparison.}
        \label{fig:intro_b}
    \end{subfigure}

    \caption{
    Overview and main benefit of HB-SJD.
    \textbf{(a)} HB-SJD replaces the token-by-token autoregressive student rollout with parallel proposal, verification, and multi-token commitment.
    \textbf{(b)} HB-SJD reduces rollout and end-to-end training time while maintaining comparable generation quality.
    }
    \label{fig:intro}
\end{figure}

However, this closer match between training and inference comes at a clear cost: OPD moves autoregressive generation directly into the training loop. Before teacher--student scoring can begin, the current student must first generate the rollout suffix. Even with KV caching, generating an $L$-token suffix still requires $L$ sequential decoding steps. As illustrated in Fig.~\ref{fig:intro}(a), this rollout lies directly between the real prefix and the completed student trajectory, making it part of the training critical path. The same bottleneck appears in GKD and recent visual autoregressive distillation methods such as VarKD \citep{agarwal2024policy,peruzzo2026knowledge}. This creates a key tension: OPD benefits from trajectories generated by the current student, but generating these trajectories is itself expensive. Can we accelerate student rollout without changing the learning process of OPD?

To address this question, we turn to Speculative Jacobi Decoding (SJD) as the student rollout engine for visual on-policy distillation. Recent work has also explored speculative decoding for online RL rollouts, showing its potential to reduce rollout cost during training \citep{iso2026accelerating}. Unlike these language-model RL settings, visual OPD repeatedly generates large batches of image trajectories from the current student. SJD is well suited to this setting because it reuses predictions from the same student as draft tokens and requires no separately trained draft model \citep{teng2025accelerating}. However, the original SJD is designed for single-sequence inference. Running it separately for each image loses batch parallelism, while processing all images in lockstep causes faster images to wait for slower ones. Efficiently adapting SJD to large-batch visual OPD is therefore the main challenge.

We therefore propose \textbf{Hybrid Batched Speculative Jacobi Decoding (HB-SJD)}, a batched SJD rollout engine for visual OPD. HB-SJD allows each image to advance according to its own decoding progress while still verifying images at different sequence positions in batched student forwards. It further uses hybrid full and compact execution to keep later rollout rounds efficient as images finish. As illustrated in Fig.~\ref{fig:intro}(a), HB-SJD only replaces the student rollout backend, while the remaining OPD training procedure is unchanged. Fig.~\ref{fig:intro}(b) shows that this replacement substantially reduces both rollout and end-to-end training time while maintaining comparable generation quality.

Our contributions are summarized as follows:

\begin{itemize}
    \item We introduce Speculative Jacobi Decoding as a rollout engine for visual on-policy distillation. By reusing predictions from the current student, SJD enables multi-token rollout without requiring an auxiliary draft model.
    
    \item We propose HB-SJD to extend SJD from single-sequence inference to large-batch visual OPD. HB-SJD combines independent per-image progress with efficient batched execution, allowing different images to advance independently while retaining GPU parallelism.
    
    \item We integrate HB-SJD as a drop-in rollout backend for existing visual OPD methods without changing their distillation objectives or training procedures. Experiments show that HB-SJD substantially reduces rollout and end-to-end training time while maintaining comparable generation quality.
\end{itemize}

\section{Related Work} 
\subsection{Visual Autoregressive Distillation} 
Knowledge distillation transfers knowledge from a large teacher to a smaller student through distribution-level or sequence-level supervision \citep{hinton2015distilling,kim2016sequence}. For generative models, recent methods further study sequence-level matching and training on student-generated trajectories. MiniLLM improves sequence-level distillation for language models, while GKD introduces on-policy training with samples generated by the current student \citep{gu2024minillm,agarwal2024policy}. Meanwhile, visual autoregressive models such as VAR and LlamaGen have shown strong scalability with next-scale and next-token image generation \citep{tian2024visual,sun2024autoregressive}. VarKD further studies supervised, sequence-level, and on-policy distillation for visual autoregressive models \citep{peruzzo2026knowledge}. In visual OPD, student-generated trajectories become part of the training loop, making rollout efficiency an important practical factor in addition to the distillation objective itself. These works mainly focus on what supervision is used and how the student is optimized. Our work is complementary: we keep the distillation formulation unchanged and focus on reducing the cost of generating student trajectories during visual on-policy distillation.

\subsection{Autoregressive Generation Acceleration} 
Speculative decoding reduces the sequential cost of autoregressive generation by proposing multiple future tokens and verifying them in parallel \citep{leviathan2023fast,chen2023speculative}. Similar ideas have been extended to visual autoregressive generation. ZipAR exploits spatial structure, Continuous Speculative Decoding studies continuous visual distributions, and LANTERN, LANTERN++, and GSD use relaxed verification to improve acceleration for image generation \citep{he2024zipar,wang2026continuous,jang2025lantern,park2025lanternpp,so2025gsd}. Speculative Jacobi Decoding (SJD) takes a different approach: it uses predictions from previous Jacobi iterations as proposals and therefore does not require an auxiliary draft model \citep{teng2025accelerating}. Following SJD, recent variants further improve proposal reuse, stability, verification, and continuation \citep{teng2025sjdpp,so2026scd,teng2025sjd2,shan2026sjdvp, zhang2026sjdsv,yu2026sjdpv,kang2026sjdpac}. Other Jacobi-based methods explore additional spatial or path-level parallelism, such as Parallel Jacobi Decoding and PathRelax \citep{liao2026pjd,lei2026pathrelax}.

Most of these methods focus on inference-time generation. Batched speculative decoding has also been studied for inference serving, where different acceptance lengths create different decoding progress across requests \citep{zhang2025batchspec}. More recently, speculative decoding has been used to reduce online rollout cost in language-model RL training \citep{iso2026accelerating}. Our setting is different: visual OPD requires the current student to repeatedly generate large batches of image trajectories during training. We therefore adapt SJD to this setting with independent per-image progress and efficient batched execution, while leaving the teacher and distillation objective unchanged.

\section{Preliminaries}
\label{sec:preliminaries}

\subsection{Visual On-Policy Distillation}
\label{sec:visual_opd}

Let $\mathbf{x}_{1:T}=(x_1,\ldots,x_T)$ denote a tokenized image. A visual autoregressive model factorizes its distribution as
\begin{equation}
p_{\theta}(\mathbf{x}_{1:T})=\prod_{t=1}^{T}p_{\theta}(x_t\mid\mathbf{x}_{<t}).
\label{eq:visual_ar_factorization}
\end{equation}
We consider a fixed teacher $p_{\mathrm{T}}$ and a student $p_{\mathrm{S},\theta}$ defined over the same visual vocabulary \citep{agarwal2024policy,peruzzo2026knowledge}.

In visual on-policy distillation, part of the sequence is generated by the current student rather than from real data or the teacher. Given a real prefix $\mathbf{x}_{1:k}^{\mathrm{real}}$, the student generates the suffix:
\begin{equation}
p_{\mathrm{mix},\theta}^{k}(\tilde{\mathbf{x}}_{1:T})
=
p_{\mathrm{data}}(\tilde{\mathbf{x}}_{1:k})
\prod_{t=k+1}^{T}
p_{\mathrm{S},\theta}(\tilde{x}_t\mid\tilde{\mathbf{x}}_{<t}).
\label{eq:mixed_student_rollout}
\end{equation}
The completed sequence is then used as context for teacher--student distillation. Following our training setting, we use bidirectional KL divergence between the teacher and student distributions:
\begin{equation}
\mathcal{L}_{\mathrm{OPD}}
=
\mathbb{E}_{\tilde{\mathbf{x}}\sim p_{\mathrm{mix},\theta}^{k}}
\left[
\frac{1}{T-k}\sum_{t=k+1}^{T}
\mathcal{D}_{\mathrm{biKL}}
\left(
p_{\mathrm{S},\theta}(\cdot\mid\tilde{\mathbf{x}}_{<t}),
p_{\mathrm{T}}(\cdot\mid\tilde{\mathbf{x}}_{<t})
\right)
\right],
\label{eq:visual_opd_objective}
\end{equation}
where $\mathcal{D}_{\mathrm{biKL}}(p,q)=\frac{1}{2}\mathrm{KL}(p\|q)+\frac{1}{2}\mathrm{KL}(q\|p)$. Since the student suffix must be generated online before distillation can proceed, conventional autoregressive rollout introduces $T-k$ sequential decoding steps into the training critical path.

\subsection{Speculative Jacobi Decoding}
\label{sec:sjd_preliminaries}

Speculative decoding reduces autoregressive decoding steps by proposing multiple future tokens and verifying them in parallel \citep{leviathan2023fast,chen2023speculative}. Speculative Jacobi Decoding (SJD) removes the auxiliary draft model by reusing predictions from previous Jacobi iterations as draft tokens \citep{teng2025accelerating}. In each iteration, a Jacobi window of draft tokens is evaluated in one model forward and verified from left to right. Accepted tokens are appended to the generated sequence, while the remaining tokens are refined for the next iteration. By committing multiple tokens in some iterations, SJD reduces the number of sequential model forwards required for autoregressive generation. The original SJD is designed for single-sequence inference, while our work extends it to batched visual OPD rollouts.

\section{Method}
\label{sec:method}

\subsection{Motivation Analysis}
\label{sec:motivation}

\begin{figure*}[t]

    \centering
    \begin{minipage}[t]{0.48\textwidth}
        \centering
        \includegraphics[
            width=\linewidth,
            trim=4pt 4pt 4pt 4pt,
            clip
        ]{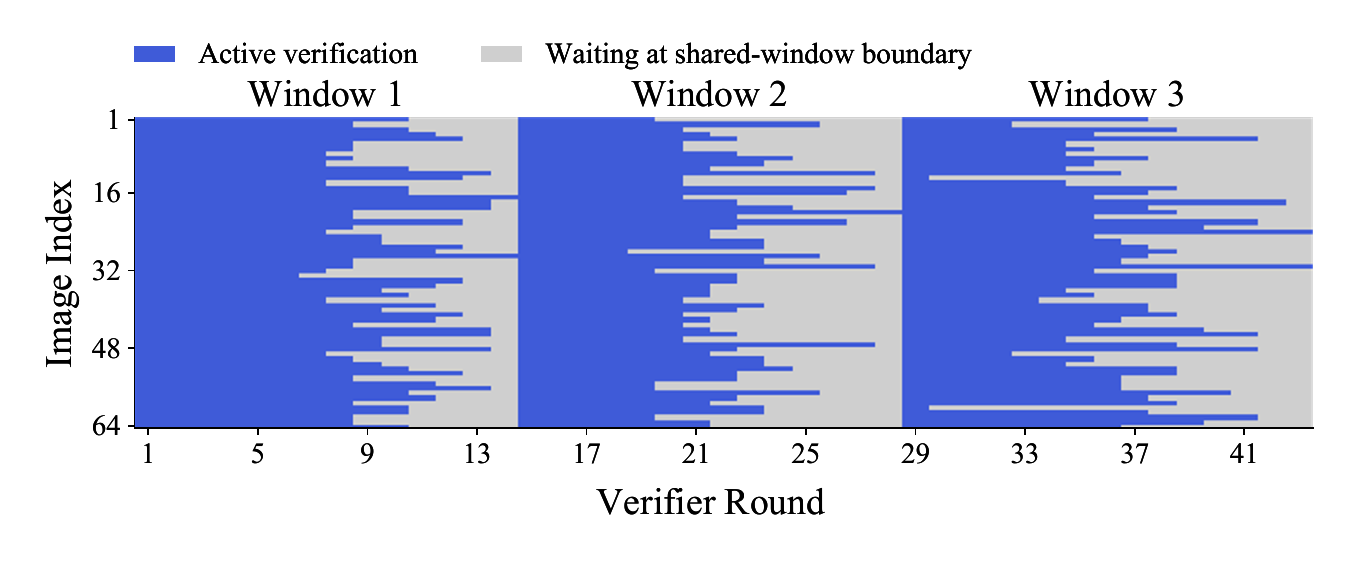}
        
        \vspace{2pt}
        \textbf{(a)} Batch-Synchronous Progress
    \end{minipage}\hfill
    \begin{minipage}[t]{0.48\textwidth}
        \centering
        \includegraphics[
            width=\linewidth,
            trim=4pt 4pt 4pt 4pt,
            clip
        ]{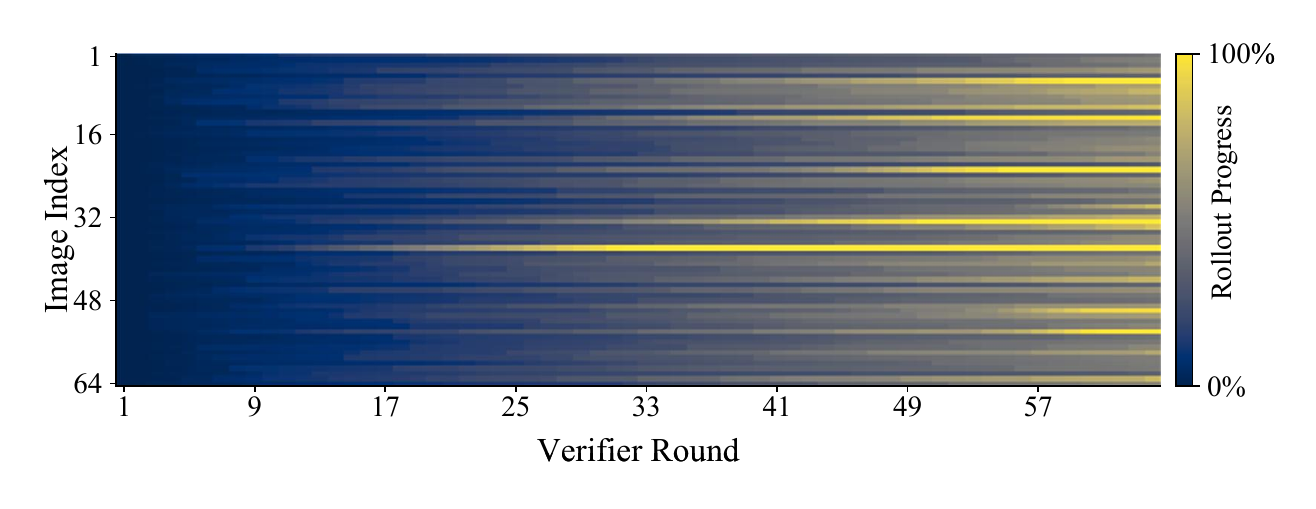}
        
        \vspace{2pt}
        \textbf{(b)} Independent Progress
    \end{minipage}
    \caption{
    Empirical motivation for independent progress.
    \textbf{(a)} Under batch-synchronous execution, faster images must wait at shared window boundaries.
    \textbf{(b)} Independent progress removes this waiting, but the number of active images gradually decreases as rollout proceeds.
    }
    \label{fig:motivation_ab}
\end{figure*}
\paragraph{Why SJD for Visual OPD.}
Visual OPD repeatedly generates trajectories from the current student, whose policy continues to change during training \citep{agarwal2024policy,peruzzo2026knowledge}. Speculative decoding methods that rely on an auxiliary draft model therefore need to keep the drafter aligned with the evolving student, introducing additional synchronization or adaptation cost \citep{iso2026accelerating}. SJD avoids this issue by reusing predictions from Jacobi iterations of the same student and requires no separately trained draft model \citep{teng2025accelerating}. These properties make SJD particularly suitable for accelerating student rollouts in visual OPD.

\paragraph{From Single-Sequence to Batched SJD.}
The original SJD is designed for single-sequence inference. Running it separately for each image would break a large rollout batch into many small model calls and lose GPU parallelism. A simple alternative, which we refer to as \emph{Batch-Synchronous SJD}, is to process all images synchronously. However, different images require different numbers of verification rounds to complete the same Jacobi window. As shown by the measured traces in Fig.~\ref{fig:motivation_ab}(a), images that finish earlier must wait for the slowest ones before the batch can move to the next window. Fig.~\ref{fig:motivation_ab}(b) shows that allowing each image to keep its own progress removes this shared-window waiting. We therefore adopt \emph{Independent Batched SJD}, where each image advances according to its own decoding progress while different images are still verified together in batched model forwards.

\vspace{3pt}
\noindent
\begin{minipage}[t]{0.61\linewidth}
\vspace{0pt}

\paragraph{Efficient Execution as the Batch Shrinks.}
Independent progress creates another problem: images finish at different times, so the number of active images gradually decreases. Removing finished images immediately seems natural, but Compact execution is not always faster because indexed KV-cache operations introduce additional overhead. As shown in Fig.~\ref{fig:motivation_c}, Full execution is more efficient when most images are still active, while Compact execution becomes preferable after the active set becomes sufficiently small. We therefore keep finished images logically inactive and switch the physical execution between Full and Compact forms according to a hardware-calibrated threshold.

\end{minipage}%
\hfill
\begin{minipage}[t]{0.36\linewidth}
\vspace{0pt}
\centering

\includegraphics[
    width=\linewidth,
    trim=5pt 5pt 5pt 5pt,
    clip
]{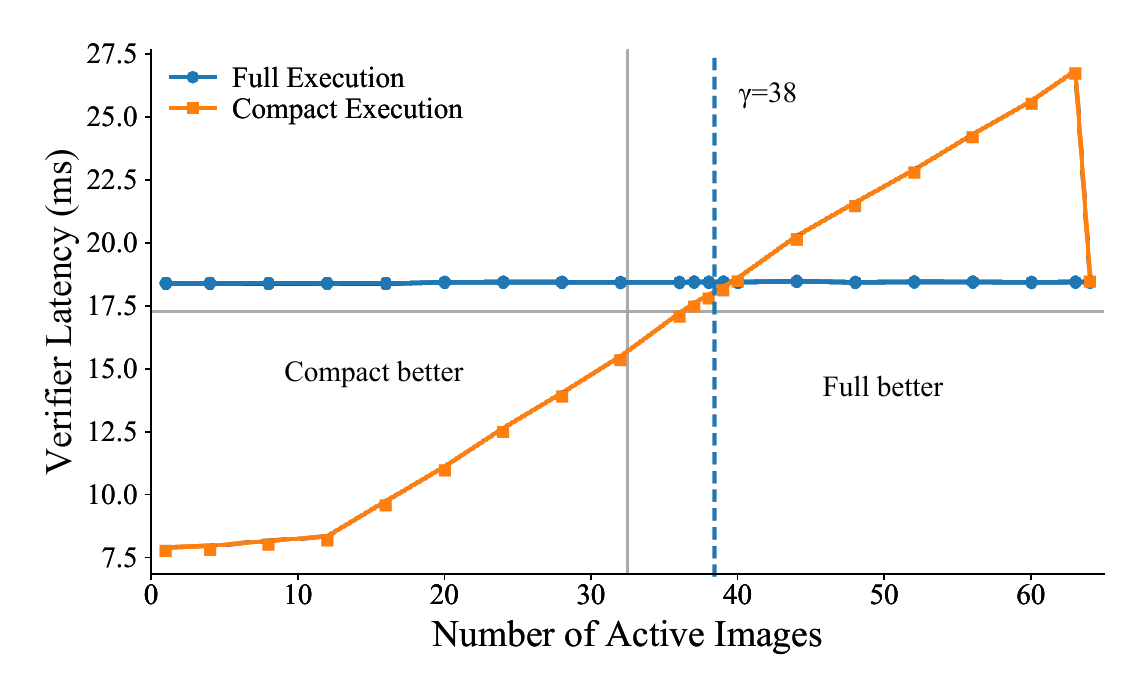}

\vspace{2pt}

\captionof{figure}{
Median verifier latency of Full and Compact execution as the number of active images decreases.
}
\label{fig:motivation_c}

\end{minipage}

\vspace{4pt}

Based on these observations, we propose \textbf{Hybrid Batched Speculative Jacobi Decoding (HB-SJD)}. HB-SJD combines independent per-image progress with efficient batched execution, extending SJD from single-sequence inference to large-batch visual OPD rollouts. It further adapts the physical execution strategy as the number of active images decreases, allowing the rollout to remain efficient throughout generation. It only replaces the student rollout backend, while the teacher, distillation objective, and optimization procedure remain unchanged.

\subsection{HB-SJD Rollout}
\label{sec:hb_sjd}

\paragraph{Independent rollout states.} At each on-policy step, the student is fixed while the rollout batch is generated. Given a real prefix, each image maintains its own committed cursor and a Jacobi window of $W$ draft tokens. After prefix pre-filling, the first draft token is obtained from the current student. In later rounds, HB-SJD first reuses predictions from the previous verification and fills the remaining positions with previously committed tokens at a fixed history offset $H$. This history-based initialization exploits the spatial locality of visual tokens instead of relying on random initialization.

\paragraph{Batched verification.}
For each active image, HB-SJD forms a verifier input using the last committed token followed by its draft window. Although different images may have different committed cursors, per-image position indices and KV-cache locations allow them to be processed together in one batched student forward. The resulting predictions are then used to verify the draft window from left to right. We consider two verification strategies.

\paragraph{Greedy verification.}
Our main implementation uses greedy rollout. Let $p_{b,j}$ denote the distribution for the $j$-th position of image $b$, and let $z_{b,j}$ be the draft token. The draft is accepted when
\begin{equation}
z_{b,j}=\arg\max_{v\in\mathcal{V}}p_{b,j}(v).
\label{eq:greedy_verification}
\end{equation}
Verification stops at the first mismatch. All matching drafts before this position are committed, and the current student argmax is committed as the replacement token. Predictions after the mismatch are not committed and are instead reused as drafts in the next round.

\paragraph{Probabilistic SJD verification.}
HB-SJD also supports the probabilistic verification rule of SJD \citep{teng2025accelerating}. Let $q_{b,j}$ be the proposal distribution associated with draft token $z_{b,j}$. Instead of requiring the draft to equal the greedy prediction, SJD accepts it with probability
\begin{equation}
\alpha_{b,j}
=
\min\left(
1,
\frac{p_{b,j}(z_{b,j})}
     {q_{b,j}(z_{b,j})}
\right).
\label{eq:sjd_verification}
\end{equation}
If a draft is rejected, the token is resampled from the calibrated distribution proportional to $[p_{b,j}-q_{b,j}]_+$, following SJD. Unlike greedy argmax matching, this criterion may accept non-argmax proposals or reject argmax proposals according to the $p/q$ ratio.

\paragraph{Independent progress.}
After verification, each image commits its accepted prefix and updates its cursor independently. Thus, different images can advance by different numbers of tokens after the same batched forward. Under classifier-free guidance, the conditional and unconditional branches of each image share the same rollout state, and verification uses the fused guided logits. After a rejection, the replacement token is used as the starting token of the next verification round. This allows the KV cache after the rejection point to be recomputed together with the next verifier forward, without an extra cache-repair step.

\subsection{Hybrid Full/Compact Execution}
\label{sec:hybrid_execution}

Independent progress allows images to finish at different times, so the number of active images gradually decreases during rollout. A straightforward solution is to remove finished images immediately and run the verifier only on the remaining active images. However, this compact execution is not always faster because selecting a subset of KV-cache rows introduces additional indexing overhead. As shown in Fig.~\ref{fig:motivation_c}, full execution is more efficient when most images are still active, while compact execution becomes preferable after the active batch becomes sufficiently small.

HB-SJD therefore supports two execution modes. \emph{Full execution} keeps the original physical batch shape, while finished images are treated as logically inactive and do not update their rollout states. \emph{Compact execution} runs the verifier only on active images and accesses their corresponding KV-cache rows. The two modes use exactly the same rollout states and verification procedure; they differ only in which rows are physically processed.

Before training, we perform a lightweight, data-independent latency test with a few verifier forwards to determine the hardware-specific switching threshold $\gamma$, which is then fixed throughout training. At verification round $r$, let $A^{(r)}$ denote the number of active images. HB-SJD selects,
\begin{equation}
\mathcal{E}^{(r)}
=
\begin{cases}
\mathrm{Full}, & A^{(r)} > \gamma,\\
\mathrm{Compact}, & A^{(r)} \leq \gamma.
\end{cases}
\label{eq:hybrid_execution}
\end{equation}
This hybrid strategy keeps the full-batch path at high occupancy and avoids unnecessary computation when only a small number of images remain active.

\begin{table*}[t]
\centering
\caption{Main results on ImageNet. HB-SJD only replaces the student rollout backend, while the distillation objective and training setup remain unchanged.}
\label{tab:main_results}
\small
\setlength{\tabcolsep}{4.5pt}
\begin{tabular}{llcccccc}
\toprule
\multirow{2}{*}{Method}
& \multirow{2}{*}{Verification}
& \multicolumn{4}{c}{Generation Quality}
& \multicolumn{2}{c}{Rollout Efficiency} \\
\cmidrule(lr){3-6} \cmidrule(lr){7-8}
& & FID $\downarrow$
& IS $\uparrow$
& Precision $\uparrow$
& Recall $\uparrow$
& Time (s) $\downarrow$
& Speedup $\uparrow$ \\
\midrule

\textbf{LlamaGen-B}
& -- & 6.51 & 156.3 & 0.81 & 0.46 & -- & -- \\

+ KD
& -- & 4.92 & 195.6 & 0.84 & 0.45 & -- & -- \\

+ SeqKD
& -- & 5.08 & 197.9 & 0.84 & 0.44 & -- & -- \\

\cmidrule(lr){1-8}

+ GKD
& AR
& 4.96 & 193.7 & 0.85 & 0.44
& 3.98 & $1.00\times$ \\

+ GKD + HB-SJD
& Greedy
& 4.95 & 192.8 & 0.85 & 0.44
& 2.68 & $1.48\times$ \\

+ GKD + HB-SJD
& Probabilistic
& 4.97 & 193.5 & 0.85 & 0.43
& 2.72 & $1.46\times$ \\

\cmidrule(lr){1-8}

+ VarKD
& AR
& 4.73 & 200.1 & 0.85 & 0.46
& 3.92 & $1.00\times$ \\

+ VarKD + HB-SJD
& Greedy
& 4.75 & 199.4 & 0.85 & 0.44
& 2.48 & $1.58\times$ \\

+ VarKD + HB-SJD
& Probabilistic
& 4.76 & 200.0 & 0.84 & 0.45
& 2.68 & $1.46\times$ \\

\midrule

\textbf{LlamaGen-L}
& -- & 3.07 & 156.0 & 0.83 & 0.52 & -- & -- \\

+ KD
& -- & 3.02 & 244.5 & 0.82 & 0.54 & -- & -- \\

+ SeqKD
& -- & 3.10 & 258.1 & 0.83 & 0.52 & -- & -- \\

\cmidrule(lr){1-8}

+ GKD
& AR
& 2.99 & 251.6 & 0.82 & 0.54
& 9.82 & $1.00\times$ \\

+ GKD + HB-SJD
& Greedy
& 3.00 & 251.4 & 0.82 & 0.54
& 6.28 & $1.56\times$ \\

+ GKD + HB-SJD
& Probabilistic
& 2.99 & 250.9 & 0.81 & 0.55
& 6.47 & $1.51\times$ \\

\cmidrule(lr){1-8}

+ VarKD
& AR
& 2.86 & 243.3 & 0.83 & 0.54
& 9.71 & $1.00\times$ \\

+ VarKD + HB-SJD
& Greedy
& 2.88 & 243.1 & 0.82 & 0.55
& 5.88 & $1.65\times$ \\

+ VarKD + HB-SJD
& Probabilistic
& 2.87 & 243.1 & 0.83 & 0.53
& 6.30 & $1.54\times$ \\

\bottomrule
\end{tabular}
\end{table*}
\section{Experiments}

\subsection{Experimental Setup}
\label{sec:experimental_setup}

\paragraph{Training and evaluation.} We follow the LlamaGen experimental setup of VarKD \citep{peruzzo2026knowledge}, including the ImageNet training and evaluation protocol. We use the same teacher--student setting, optimization setup, and number of training steps for all compared rollout methods. Generation quality is evaluated on 50K samples using FID, Inception Score (IS), precision, and recall. Reported main latency results are averaged over three independent runs.

\paragraph{Rollout settings.} We compare HB-SJD with the KV-cached autoregressive rollout under the same training setup. We refer to this token-by-token autoregressive rollout with KV caching as \emph{Cached AR}. Unless otherwise specified, HB-SJD uses a Jacobi window size of $W=16$, a history offset of $H=24$, and classifier-free guidance with scale $2.0$. We evaluate two verification strategies: \emph{Greedy Verification}, which verifies drafts against the current student's greedy predictions, and \emph{Probabilistic Verification}, which follows the probabilistic acceptance and resampling rule of SJD \citep{teng2025accelerating}. We report rollout latency and end-to-end training time for efficiency evaluation. For latency distributions, P95 denotes the 95th-percentile latency.

\paragraph{Baselines and evaluation.}
For LlamaGen, KD, and SeqKD, we report the generation-quality results from VarKD under its ImageNet evaluation protocol. For GKD, VarKD, and their HB-SJD variants, we reproduce the results in our environment using the same training and evaluation setup. All rollout efficiency results are measured on the same NVIDIA H200 GPU under identical settings.

\subsection{Main Results}
\label{sec:main_results}

\paragraph{Quantitative results.}
Table~\ref{tab:main_results} compares HB-SJD with visual autoregressive distillation baselines on LlamaGen-B and LlamaGen-L. HB-SJD only replaces the student rollout backend and leaves the distillation objective and training procedure unchanged. With Greedy Verification, HB-SJD achieves $1.48\times$ and $1.56\times$ rollout speedups for GKD on LlamaGen-B and LlamaGen-L, respectively, while the corresponding speedups for VarKD reach $1.58\times$ and $1.65\times$. Probabilistic Verification also provides consistent acceleration, achieving $1.46$--$1.54\times$ speedups across different settings. Meanwhile, FID, IS, precision, and recall remain close to the AR rollout baselines. These results show that HB-SJD consistently reduces rollout cost across model sizes and distillation methods while maintaining comparable generation quality across all evaluated training settings.

\begin{figure}[t]
    \centering
    \includegraphics[
        width=\linewidth,
        trim=5pt 5pt 5pt 5pt,
        clip
    ]{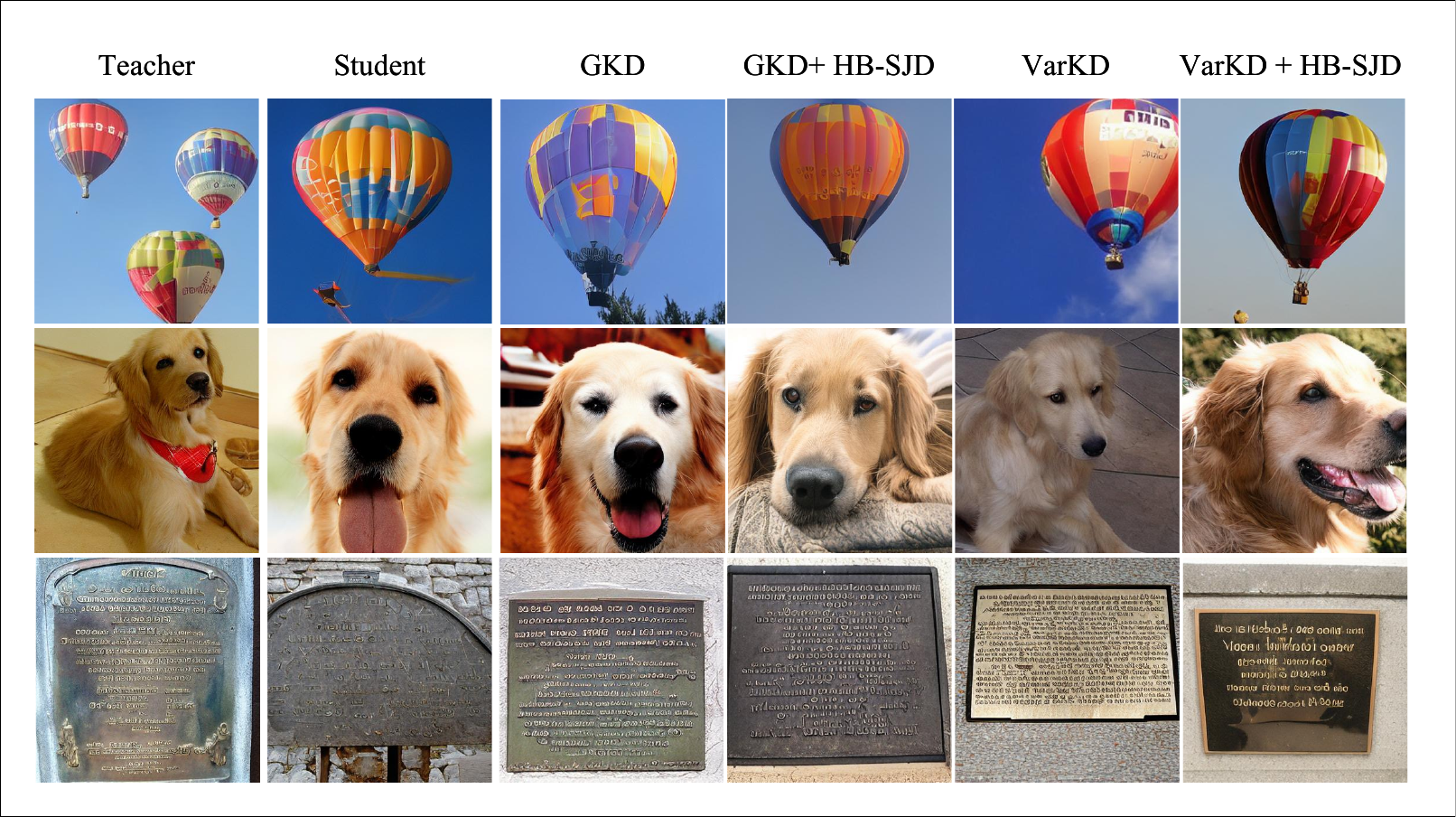}
    \caption{Qualitative comparison of generation results using Greedy Verification for HB-SJD.}
    \label{fig:quality}
\end{figure}

\paragraph{Qualitative results.}
Figure~\ref{fig:quality} shows qualitative results using LlamaGen-XL as the teacher and LlamaGen-L as the student. Compared with the corresponding GKD and VarKD baselines, models trained with HB-SJD produce visually comparable samples in semantic content, object structure, and local details. These results are consistent with the quantitative evaluation, indicating that replacing autoregressive rollout with HB-SJD does not introduce noticeable degradation in generation quality.

\subsection{Ablation Study}
\label{sec:ablation}

\paragraph{Is independent progress necessary for batched SJD?} To answer this question, we compare Cached AR, Batch-Synchronous SJD, and Independent Batched SJD under the same rollout workload. Here, Independent uses Full execution. As shown in Table~\ref{tab:batch_execution}, batch-synchronous SJD requires 241 verification rounds and takes 4239 ms, making it slightly slower than Cached AR at 3982 ms. In contrast, allowing each image to advance independently reduces the number of verification rounds to 176 and the rollout latency to 3275 ms, achieving a $1.216\times$ speedup over Cached AR. This result shows that preserving batched model execution alone is not sufficient; independent per-image progress is important for efficiently adapting SJD to large-batch visual OPD. Since the synchronous and independent implementations also differ in proposal construction, we treat this comparison as a system-level ablation rather than attributing the entire improvement to scheduling alone.

\paragraph{Is Hybrid execution necessary as the active batch shrinks?} To answer this question, we conduct a separate execution sweep comparing Full, Compact, and Hybrid execution under the same Independent Batched SJD rollout. Full execution keeps the full physical batch throughout rollout, while Compact execution only processes the remaining active images. Hybrid execution switches between them according to the calibrated threshold $\gamma$. As shown in Table~\ref{tab:hybrid_execution_ablation}, Compact execution reduces the mean latency from $3292$ ms to $3104$ ms, but a single execution mode cannot remain optimal as the active batch changes. Hybrid execution further reduces the mean latency to $2784$ ms and the P95 latency from $4714$ ms to $3969$ ms, achieving a $1.182\times$ speedup over Full execution. This corresponds to a $15.4\%$ latency reduction over Full execution and a $10.3\%$ reduction over Compact execution. These results confirm that dynamically switching the execution mode is more effective than using either Full or Compact execution throughout the entire rollout.

\paragraph{Does HB-SJD benefit more from longer rollouts?} To answer this question, we vary the prefix ratio while keeping the model, batch size, and other rollout settings unchanged. A smaller prefix ratio corresponds to a longer student-generated suffix and therefore more sequential decoding steps for Cached AR. As shown in Table~\ref{tab:rollout_length}, HB-SJD consistently accelerates rollout across all tested lengths. When the prefix ratio decreases from $0.75$ to $0.20$, the suffix length increases from $144$ to $461$ tokens, and the speedup increases from $1.392\times$ to $1.699\times$. This trend shows that HB-SJD provides larger benefits for longer rollouts, where reducing sequential autoregressive decoding becomes increasingly important.

\begin{table}[t]
\noindent
\hspace*{4pt}%
\begin{minipage}[t]{0.54\linewidth}
\vspace{0pt}
\centering
\caption{Effect of batched rollout execution.}
\label{tab:batch_execution}

{\setlength{\tabcolsep}{4pt}
\begin{tabular}[t]{@{}lccc@{}}
\hline
\textbf{Method} &
\textbf{Time (ms)$\downarrow$} &
\textbf{Speedup$\uparrow$} &
\textbf{Rounds $\downarrow$} \\
\hline
Cached AR
& 3982
& $1.000\times$
& -- \\
Sync. Batch
& 4239
& $0.939\times$
& 241 \\
Independent
& \textbf{3275}
& \textbf{$1.216\times$}
& \textbf{176} \\
\hline
\end{tabular}
}
\end{minipage}%
\hfill
\begin{minipage}[t]{0.44\linewidth}
\vspace{0pt}
\centering
\caption{Effect of hybrid execution.}
\label{tab:hybrid_execution_ablation}

{\setlength{\tabcolsep}{1.5pt}
\begin{tabular}[t]{@{}lccc@{}}
\hline
\textbf{Mode} &
\textbf{Mean (ms)} &
\textbf{P95 (ms)} &
\textbf{vs. Full $\uparrow$} \\
\hline
Full
& 3292
& 4714
& $1.000\times$ \\
Compact
& 3104
& 4339
& $1.061\times$ \\
Hybrid
& \textbf{2784}
& \textbf{3969}
& \textbf{$1.182\times$} \\
\hline
\end{tabular}
}
\end{minipage}

\end{table}
\paragraph{Does HB-SJD remain effective across different rollout batch sizes?} To answer this question, we vary the rollout batch size from $8$ to $64$ while keeping the model and other rollout settings unchanged. For each batch size, HB-SJD uses the corresponding hardware-calibrated execution setting. As shown in Table~\ref{tab:batch_size}, HB-SJD consistently outperforms Cached AR across all tested batch sizes, achieving speedups from $1.298\times$ to $1.486\times$. In particular, HB-SJD achieves $1.403\times$ and $1.298\times$ speedups even at batch sizes $8$ and $16$, while maintaining $1.486\times$ and $1.483\times$ speedups at batch sizes $32$ and $64$. These results show that the acceleration of HB-SJD is not limited to a specific rollout batch size and remains effective under different batch workloads.

\begin{table}[t]
\centering

\begin{minipage}[t]{0.49\linewidth}

\centering
\caption{Effect of rollout length.}
\label{tab:rollout_length}

\begin{tabular*}{\linewidth}[t]{@{\extracolsep{\fill}}ccccc@{}}
\hline
\noalign{\vskip 1pt}
\textbf{\shortstack{Prefix\\Ratio}} &
\textbf{Suffix} &
\textbf{\shortstack{AR\\(ms)}} &
\textbf{\shortstack{HB-SJD\\(ms)}} &
\textbf{Speedup $\uparrow$} \\
\hline
0.20 & 461 & 6363 & 3745 & \textbf{$1.699\times$} \\
0.40 & 346 & 4803 & 3101 & $1.549\times$ \\
0.60 & 231 & 3242 & 2202 & $1.472\times$ \\
0.75 & 144 & 2061 & 1481 & $1.392\times$ \\
\hline
\end{tabular*}

\end{minipage}%
\hfill
\begin{minipage}[t]{0.49\linewidth}
\centering
\caption{Effect of rollout batch size.}
\label{tab:batch_size}

\begin{tabular*}{\linewidth}[t]{@{\extracolsep{\fill}}cccc@{}}
\hline
\noalign{\vskip 1pt}
\textbf{$B$} &
\textbf{\shortstack{AR\\(ms)}} &
\textbf{\shortstack{HB-SJD\\(ms)}} &
\textbf{Speedup $\uparrow$} \\
\hline
8  & 1737 & 1238 & $1.403\times$ \\
16 & 1781 & 1372 & $1.298\times$ \\
32 & 2520 & 1696 & \textbf{$1.486\times$} \\
64 & 3980 & 2685 & $1.483\times$ \\
\hline
\end{tabular*}

\end{minipage}

\end{table}

\paragraph{Does a larger Jacobi window always improve rollout efficiency?} To answer this question, we vary the Jacobi window size $W$ under Full execution while keeping other rollout settings unchanged. A larger window allows the verifier to process more draft tokens in each round and increases the number of committed tokens. As shown in Table~\ref{tab:window_size}, the mean number of committed tokens increases from $1.746$ at $W=4$ to $2.951$ at $W=24$. However, a larger window also increases the cost of each verifier forward, so rollout latency is not monotonic in $W$. $W=8$ achieves the lowest latency of $2912$ ms and a $1.384\times$ speedup, while further increasing $W$ provides higher commit rates but lower end-to-end efficiency. We use $W=16$ as the fixed setting in our full training experiments, while this ablation highlights the trade-off between per-round progress and verifier cost.

\begin{table}[t]
\centering

\begin{minipage}[t]{0.49\linewidth}
\vspace{0pt}
\centering
\caption{Effect of Jacobi window size.}
\label{tab:window_size}
\begin{tabular}[t]{cccc}
\hline
\textbf{$W$} & \textbf{Time (ms) $\downarrow$} & \textbf{Speedup $\uparrow$} & \textbf{Commit $\uparrow$} \\
\hline
4 & 3449 & $1.169\times$ & 1.746 \\
8 & \textbf{2912} & \textbf{$1.384\times$} & 2.162 \\
12 & 3067 & $1.315\times$ & 2.457 \\
16 & 3305 & $1.220\times$ & 2.641 \\
24 & 3751 & $1.075\times$ & \textbf{2.951} \\
\hline
\end{tabular}
\end{minipage}
\hfill
\begin{minipage}[t]{0.49\linewidth}
\vspace{0pt}
\centering
\caption{Sensitivity to history offset.}
\label{tab:history_offset}
\begin{tabular}[t]{cccc}
\hline
\textbf{$H$} & \textbf{Time (ms) $\downarrow$} & \textbf{Speedup $\uparrow$} & \textbf{Commit $\uparrow$} \\
\hline
16 & 3266 & $1.227\times$ & 2.592 \\
20 & 3320 & $1.207\times$ & 2.598 \\
24 & 3288 & $1.219\times$ & \textbf{2.641} \\
28 & 3392 & $1.181\times$ & 2.571 \\
32 & \textbf{3258} & \textbf{$1.230\times$} & 2.554 \\
\hline
\end{tabular}
\end{minipage}

\end{table}

\begin{table}[t]
\centering

\begin{minipage}[t]{0.44\linewidth}
\vspace{0pt}
\centering
\caption{Sensitivity to switching threshold.}
\label{tab:threshold}
\begin{tabular}[t]{ccc}
\hline
\textbf{$\gamma$} & \textbf{Time (ms) $\downarrow$} & \textbf{vs. Full $\uparrow$} \\
\hline
30 & 2786 & $1.180\times$ \\
38 & 2794 & $1.176\times$ \\
46 & 2792 & $1.177\times$ \\
\hline
\end{tabular}
\end{minipage}
\hfill
\begin{minipage}[t]{0.51\linewidth}
\vspace{0pt}
\centering
\caption{Efficiency at different training stages.}
\label{tab:training_stage}
\begin{tabular}[t]{cccc}
\hline
\textbf{Step} & \textbf{AR} (ms) & \textbf{HB-SJD} (ms)& \textbf{Speedup $\uparrow$} \\
\hline
2K & 4005 & 2617 & \textbf{$1.530\times$} \\
12K & 4007 & 2672 & $1.500\times$ \\
22K & 3980 & 2685 & $1.483\times$ \\
\hline
\end{tabular}
\end{minipage}

\end{table}

\begin{table}[!h]
\centering
\caption{Comparison of verification strategies.}
\label{tab:verification_strategy}
{\setlength{\tabcolsep}{5pt}
\begin{tabular}{lcccc}
\hline
\textbf{Verification} &
\textbf{Speedup $\uparrow$} &
\textbf{Reject $\downarrow$} &
\textbf{Commit $\uparrow$} &
\textbf{Forwards $\downarrow$} \\
\hline
Greedy
& \textbf{$1.613\times$}
& \textbf{37.4\%}
& \textbf{2.622}
& 162.0 \\
Probabilistic
& $1.500\times$
& 43.4\%
& 2.296
& \textbf{157.7} \\
\hline
\end{tabular}}
\end{table}

\paragraph{Is HB-SJD sensitive to the history offset $H$?} To answer this question, we vary $H$ under Full execution while keeping the other rollout settings unchanged. As shown in Table~\ref{tab:history_offset}, the rollout latency remains relatively stable across $H=16$ to $32$, with speedups ranging from $1.181\times$ to $1.230\times$. Although $H=32$ gives the lowest latency in this sweep, it improves over the default $H=24$ by less than $1\%$. These results show that HB-SJD is not sensitive to the history offset.

\paragraph{Does HB-SJD require precise tuning of the switching threshold $\gamma$?} To answer this question, we conduct a threshold sweep around the calibrated value while keeping settings unchanged. As shown in Table~\ref{tab:threshold}, the latency remains similar for $\gamma=30$, $38$, and $46$, with speedups of $1.180\times$, $1.176\times$, and $1.177\times$ over Full execution, respectively. The variation shows that Hybrid execution is insensitive to the switching point. This supports our data-independent hardware calibration instead of requiring careful tuning of $\gamma$ for each training run.

\paragraph{Does HB-SJD remain effective throughout OPD training?} To answer this question, we evaluate rollout efficiency using student checkpoints from different training stages. As shown in Table~\ref{tab:training_stage}, HB-SJD achieves $1.530\times$, $1.500\times$, and $1.483\times$ speedups at 2K, 12K, and 22K steps, respectively. Although the speedup varies slightly as the student changes, HB-SJD consistently outperforms Cached AR at all three stages. This result shows that the rollout acceleration remains effective throughout OPD training rather than depending on a particular student checkpoint.

\paragraph{How do Greedy and Probabilistic Verification differ?} To answer this question, we compare the two verification strategies in a dedicated same-checkpoint benchmark. Table~\ref{tab:verification_strategy} reports averages over LlamaGen-B and LlamaGen-L. The result shows that Greedy Verification achieves a higher average speedup of $1.613\times$, compared with $1.500\times$ for Probabilistic Verification. Although Probabilistic Verification uses slightly fewer verifier forwards, it has a higher rejection rate and a lower mean commit length. It also requires full-vocabulary probability computation and rejection resampling, resulting in higher overall rollout cost. Nevertheless, Probabilistic Verification follows the standard SJD probabilistic verification and resampling procedure, making it compatible with future SJD optimizations that may further accelerate rollout.

\section{Conclusion} 
We presented HB-SJD, an efficient rollout engine for visual on-policy distillation. HB-SJD extends Speculative Jacobi Decoding from single-sequence inference to large-batch training by allowing each image to maintain independent decoding progress while preserving batched student verification. It further combines Full and Compact execution according to a lightweight hardware-calibrated threshold, reducing both synchronization overhead and later-stage rollout cost. Importantly, HB-SJD only replaces the student rollout backend and leaves the teacher, distillation objective, and optimization procedure unchanged. Experiments on visual autoregressive models show that HB-SJD substantially reduces rollout and end-to-end training cost while maintaining comparable generation quality, with consistent acceleration across different rollout lengths, batch sizes, and training stages. 

\FloatBarrier
\bibliography{iclr2027_conference}
\bibliographystyle{iclr2027_conference}

\end{document}